\documentclass[preprint,12pt]{elsarticle}
\usepackage{amsmath}
\usepackage{amssymb}
\usepackage{algorithm} 
\usepackage{algpseudocode} 
\usepackage[utf8]{inputenc}
\usepackage[T1]{fontenc}
\usepackage{graphicx}
\usepackage{amsthm}
\usepackage{xcolor}
\usepackage{soul}
\usepackage{verbatim}

\usepackage{amssymb}
\journal{Information Sciences}

\begin{document}

\begin{frontmatter}



\title{Constrained Hybrid Metaheuristic
Algorithm for Probabilistic Neural Networks Learning}

\author[add1,add2]{Piotr A. Kowalski \fnref{fn1}}
 \ead{pkowal@agh.edu.pl}\ead{pakowal@ibspan.waw.pl}
\author[add3]{Szymon Kucharczyk \fnref{fn2}}
 \ead{kucharcz@agh.edu.pl}
\author[add4,add5]{Jacek Ma{\'n}dziuk \fnref{fn3}}
\ead{jacek.mandziuk@pw.edu.pl}

\address[add1]{
Faculty of Physics and Applied Computer Science, %\\
AGH University of Krakow,\\
al. A. Mickiewicza 30, 30-059 Cracow, Poland.
}
\address[add2]{
Systems Research Institute, %\\
Polish Academy of Sciences,\\
ul. Newelska 6, 01-447 Warsaw, Poland.
}
\address[add3]{
AGH Doctoral School, %\\
AGH University of Krakow,\\
al. A. Mickiewicza 30, 30-059 Cracow, Poland.
}
\address[add4]{
Faculty of Mathematics and Information Science, %\\
Warsaw University of Technology \\
Koszykowa 75, 00-662 Warsaw, Poland.
}
\address[add5]{
Faculty of Computer Science, %\\
AGH University of Krakow,\\
al. A. Mickiewicza 30, 30-059 Cracow, Poland.
}

\fntext[fn1]{0000-0003-4041-6900}
\fntext[fn2]{0009-0002-2413-6984} 
\fntext[fn3]{0000-0003-0947-028X}

\begin{abstract}
This study investigates the potential of hybrid metaheuristic algorithms to enhance the training of Probabilistic Neural Networks (PNNs) by leveraging the complementary strengths of multiple optimisation strategies. Traditional learning methods, such as gradient-based approaches, often struggle to optimise high-dimensional and uncertain environments, while single-method metaheuristics may fail to exploit the solution space fully. To address these challenges, we propose the constrained Hybrid Metaheuristic (cHM) algorithm, a novel approach that combines multiple population-based optimisation techniques into a unified framework. The proposed procedure operates in two phases: an initial probing phase evaluates multiple metaheuristics to identify the best-performing one based on the error rate, followed by a fitting phase where the selected metaheuristic refines the PNN to achieve optimal smoothing parameters. This iterative process ensures efficient exploration and convergence, enhancing the network's generalisation and classification accuracy.
%The hybrid method 
cHM integrates several popular metaheuristics, such as BAT, Simulated Annealing, Flower Pollination Algorithm, Bacterial Foraging Optimization, and Particle Swarm Optimisation as internal optimisers. To evaluate 
%the proposed approach's 
cHM performance, experiments were conducted on 16 datasets with varying characteristics, including binary and multiclass classification tasks, balanced and imbalanced class distributions, and diverse feature dimensions. 
The results demonstrate that 
%the constrained Hybrid Metaheuristic algorithm 
cHM effectively combines the strengths of individual metaheuristics, 
%enabling 
leading to faster convergence and more robust learning. By optimising the smoothing parameters of PNNs, the proposed method enhances classification performance across diverse datasets, proving its 
%adaptability 
application flexibility and efficiency. 
\end{abstract}


\begin{highlights}
\item Proposes a hybrid metaheuristic algorithm to optimise Probabilistic Neural Networks. 
\item Combines swarm-based methods to improve convergence and classification accuracy. 
\item Adaptively optimises smoothing parameters with an efficient exploration of solution spaces. 
\item Validates the algorithm using 16 datasets with diverse class distributions and features. 
\item Ensures computational efficiency through constrained and reproducible training processes.
\end{highlights}


\end{frontmatter}


\section{Introduction}
\label{sec:intro}
Artificial Intelligence (AI) has rapidly evolved from a niche field of academic research to an integral part of everyday life, revolutionising industries and shaping the way we interact with technology~\cite{lemonde2024nobel}. Initially, AI was seen as a distant ambition, confined to science fiction and theoretical discussions. However, advancements in machine learning, neural networks, and computational power have fueled its exponential growth, making AI accessible and applicable in numerous sectors. Today, AI is embedded in a wide range of technologies—from the personal assistants on our smartphones and recommendation algorithms on streaming platforms to advanced medical diagnostics and autonomous vehicles \cite{kasinidou2023artificial,yin2021role}.

Integrating AI into daily life has transformed how we work, communicate, and make decisions. It has enabled businesses to optimise operations, improve customer experiences, and innovate in previously unimaginable ways. For instance, AI-powered chatbots are now common in customer service, providing immediate responses and personalised support. In healthcare, AI is being used for early detection of diseases, analysis of medical images, and even drug discovery, saving lives and improving the quality of care~\cite{hamdan2021applications,bajwa2021artificial}. In the financial sector, AI models are employed to predict market trends, manage risks, and detect fraud, offering new opportunities for growth and security~\cite{li2023artificial}.

As AI continues to advance, its impact will only deepen, offering solutions to some of society’s most pressing challenges, such as climate change, education, and resource management~\cite{olawade2024artificial}, as well as public safety~\cite{KarwowskiMandziuk2019,Karwowskietal2019} and protection of natural resources~\cite{Fangetal2015,ijcai2022p88}. However, this rapid expansion also raises concerns regarding ethical implications, job displacement, and the potential for bias in decision-making systems. The future of AI will require careful consideration of these challenges, ensuring that its benefits are harnessed responsibly and equitably~\cite{gillies2022can}. Despite these concerns, the relentless progress of AI promises to reshape industries and societies, creating new opportunities and fundamentally altering the fabric of modern life~\cite{wu2022sustainable}.

AI has become a powerful force driving innovation across industries, largely due to the development of neural networks \cite{lecun2015deep}. These networks, particularly deep learning models, have revolutionised AI by enabling machines to learn from vast amounts of data and make decisions with impressive accuracy \cite{krizhevsky2012imagenet}. Neural networks are capable of identifying complex patterns in data, which has made them essential in tasks like image recognition, natural language processing, and autonomous systems \cite{vaswani2017attention}. Their ability to process and adapt to high-dimensional data has transformed fields such as healthcare, finance, and robotics \cite{esteva2017dermatologist}. However, as AI continues to grow, challenges around computational demands, interpretability, and ethical concerns remain \cite{lipton2018mythos,amodei2016concrete}. Despite these hurdles, neural networks are at the heart of AI's evolution, shaping the future of technology and offering immense potential for innovation and societal advancement \cite{goodfellow2020deep}.

One of the key branches of neural networks is the group of Probabilistic Neural Networks, which have gained significant importance in recent years due to their ability to model uncertainty and make probabilistic predictions. PNNs are designed to estimate the probability distribution of data, allowing them to provide not just predictions but also a measure of confidence in those predictions. This makes them particularly useful in applications where understanding the uncertainty or variability in the data is crucial, such as in medical diagnostics, risk assessment, and decision-making under uncertainty.

The development of PNN's can be traced back to the need for more interpretable and robust models in fields that require high levels of certainty, such as finance and healthcare. Unlike traditional neural networks, which output point estimates, PNNs generate probability distributions, offering a richer and more nuanced understanding of the data. This capability has made them invaluable in situations where the costs of misclassification are high, and decision-makers need to assess not only the most likely outcome but also the associated risks.

As research in probabilistic modelling advanced, PNNs evolved to incorporate more sophisticated algorithms for density estimation, classifying patterns with greater accuracy even in noisy or incomplete datasets. These networks have become increasingly popular in areas like pattern recognition, anomaly detection, and classification tasks where uncertainty is inherent. The significance of PNNs lies in their ability to combine the power of neural networks with the flexibility and interpretability of probabilistic models, allowing for more informed, data-driven decisions in complex, uncertain environments.

PNNs differ from traditional feedforward networks in that they do not possess the classic dense layers commonly found in such networks, nor do they rely on gradient-based learning procedures \cite{pollard2012convergence}. Instead, PNNs are typically based on kernel density estimation, where each neuron represents a local probability distribution rather than a specific learned weight \cite{HE2018374}. This structure allows PNNs to effectively model uncertainty and complex distributions, but it also means that they do not use the backpropagation algorithm or other gradient-based optimisation methods for training \cite{kusy2015application}.

The process of training PNNs, particularly in determining smoothing parameters, can be categorized into two main groups of methods. 
First, statistical deterministic approach methods, such as cross-validation \cite{silverman2018density} and plug-in techniques \cite{wand1994multivariate}. These methods focus on optimizing the smoothing parameter to minimize the error between the estimated and true probability density functions. They are stable and consistent, making them ideal when statistical accuracy is the primary goal. On the other hand, non-deterministic metaheuristic approaches, including optimization algorithms like Genetic Algorithms, Particle Swarm Optimization \cite{kowalski2020probabilistic}, and reinforcement learning \cite{kusy2015application}, prioritize classification performance by maximizing class separability. They are less stable than statistical methods but can achieve significantly better results in specific runs when the training algorithm is appropriately tuned. The method proposed in this article belongs to the second group. While it is less stable than statistical methods, it can discover significantly better solutions in individual training runs. Non-deterministic approaches allow for flexibility and adaptability during training. With appropriate algorithm triggering and fine-tuning, these methods can significantly outperform deterministic statistical techniques, especially in complex classification scenarios where achieving optimal class distinction is critical.

Learning in PNNs, however, remains a crucial procedure for their effectiveness. It is through this learning process that the network is able to generalise the knowledge contained in the data and make accurate predictions. In PNNs, the learning typically involves adjusting the smoothing parameters, which govern how the probability distributions are modelled, ensuring that the network can adapt to the underlying data distribution. This adaptation is vital, as it allows PNNs to generalise well to new, unseen data, making them highly effective for tasks such as classification and pattern recognition. Thus, while PNNs operate under a different paradigm than traditional neural networks, their learning process is essential for enabling the network to extract meaningful insights from data and apply this knowledge to make reliable predictions in real-world scenarios \cite{MUGDADI200449}.

Metaheuristic algorithms are a class of optimisation techniques that have gained prominence in neural network learning due to their ability to explore large, complex search spaces without relying on gradient-based methods. These algorithms are particularly useful for training neural networks with non-convex objective functions, where traditional methods like gradient descent may struggle to find global optima or may get stuck in local minima \cite{kaveh2023application}. Metaheuristics, such as Genetic Algorithms (GAs), Particle Swarm Optimization (PSO), Simulated Annealing (SA), Ant Colony Optimization (ACO), and other, provide an alternative approach by mimicking natural or biological processes to guide the search for optimal solutions in complex problems~\cite{kowalski2016training,Okulewiczetal2022,Zaborskietal2022}. In the context of neural networks, metaheuristics are employed to fine-tune network architecture, select optimal hyperparameters, and train the network in an efficient and robust manner. Their ability to balance exploration and exploitation makes them effective in scenarios where traditional learning techniques may fail, particularly in high-dimensional, noisy, or complex data environments. By adapting to the problem at hand, metaheuristics offer a versatile and powerful toolset for improving the performance and generalisation ability of neural networks \cite{abd2021advanced}.

\subsection{Motivation} The synergy between heuristic algorithms in neural network training can significantly enhance model performance by combining the strengths of different optimisation strategies~\cite{ul2022optimizing}. While metaheuristics excel at exploring the solution space and escaping from local optima, they may not always converge quickly or precisely to an optimal solution. To address this, hybrid approaches that combine metaheuristics with traditional gradient-based methods are increasingly being explored \cite{hosseini2024meta}. For example, a metaheuristic algorithm can be used to find a promising starting point or optimise the network's architecture, followed by fine-tuning the weights with a gradient descent method. Alternatively, multiple metaheuristic algorithms can be combined to leverage the diverse exploration capabilities of each, ensuring a more thorough search of the solution space. This synergy can lead to faster convergence, better global search, and improved generalisation, making it an effective strategy for complex neural network training tasks. The complementary nature of heuristic and metaheuristic algorithms allows for a more robust and adaptive learning process, ultimately enhancing the network's ability to solve various challenging problems \cite{chiroma2020nature}.

The motivation behind this research lies in the continuous quest to improve the efficiency and effectiveness of learning algorithms for neural networks, particularly in the context of Probabilistic Neural Networks (PNNs). Traditional learning methods, including gradient-based approaches, face challenges in optimising complex models, especially in high-dimensional and uncertain environments. While metaheuristic algorithms, particularly swarm-based techniques like Particle Swarm Optimization (PSO) and Firefly Algorithm (FPA), have shown promise in exploring complex solution spaces, their application is often limited to single-method optimisation processes that may not fully exploit the potential of the search space.

\subsection{Contribution} In this work, we introduce a novel approach—constrained Hybrid Metaheuristic (cHM)—that seeks to overcome these limitations by combining multiple weak metaheuristics into a more robust and efficient super-metaheuristic. By leveraging the strengths of several population-based optimisation techniques, the cHM procedure can more effectively explore and optimise the smoothing parameters for PNNs, improving the network's performance. This hybrid method is particularly valuable for PNNs, where selecting appropriate smoothing parameters plays a crucial role in generalisation and classification accuracy.

The proposed cHM procedure is designed to operate in two phases: probing and fitting, with each phase subject to time constraints to ensure computational efficiency. The probing phase allows for an initial evaluation of multiple metaheuristics, selecting the one that performs best in terms of error rate. In the fitting phase, the best-performing metaheuristic continues to refine the PNN, ensuring that the model converges to an optimal solution. This iterative process allows the system to adapt and improve with each cycle, ensuring robust learning and high-quality predictions. Through this innovative hybrid approach, we aim to enhance the learning capabilities of PNNs, providing a more efficient and flexible tool for solving complex classification tasks.

\subsection{Paper organization}
The remainder of this paper is organised as follows. Section \ref{sec:PNN} provides a detailed overview of the probabilistic neural network, including its structure and the training procedures, with a particular focus on the smoothing parameter modification procedure. Section \ref{sec:cHM} introduces the constrained Hybrid Metaheuristic algorithm, describing the various metaheuristic techniques considered, such as Particle Swarm Optimisation, the BAT algorithm, Bacterial Foraging Optimization, Simulated Annealing, and the Flower Pollination Algorithm. The proposed algorithm is then presented in detail. Section \ref{sec:Results} discusses the results of numerical investigations, covering PNN training details, data sets, and the outcomes of the proposed learning procedure, including performance analysis of the metaheuristic training methods and the frequency of metaheuristic selection. Finally, Section\ref{sec:Conclusions} concludes the paper with a summary of findings and potential future research directions.

\section{Probabilistic neural network}
\label{sec:PNN}
Probabilistic neural networks (PNNs) are a type of artificial neural network that incorporate probabilistic principles in their functioning. In 1990 Donald Specht introduced the concept of 
%probabilistic neural networks 
PNNs in his papers \cite{Specht_1990_IEEE_NN,Specht_1990_NN}, which presented a new approach to classification \cite{Mantzaris} and regression \cite{Wen} tasks by combining statistical methods with neural networks. Largely influenced by Bayes’ theorem and the Parzen window method for probability density estimation, PNNs aim to combine the strengths of statistical probability with the powerful learning capabilities of neural networks. This enables them to model uncertainty in data more effectively. Unlike traditional neural networks that generate deterministic outputs, PNNs focus on predicting a probability distribution over possible outcomes. This allows them to handle noisy or uncertain input data and provide more robust predictions in real-world applications.

There are two main types of probabilistic neural networks: those designed for regression and those intended for classification tasks. In regression-based PNNs, the goal is to predict continuous values while providing an estimate of the uncertainty associated with those predictions. These models typically return a mean prediction along with a confidence interval, enabling decision-making that considers not just the prediction but also the reliability of the output. In contrast, classification-based PNNs focus on assigning input data to discrete categories. They do this by estimating the probability that an input belongs to each possible class and then selecting the class with the highest probability. This probabilistic approach to classification helps in handling ambiguous or noisy data, as the model doesn’t just provide a single label but rather a probability distribution over all possible classes (labels).

The development of 
%probabilistic neural networks 
PNNs has been shaped by advances in both computational power and probabilistic modelling techniques. Early PNNs were limited by the computational cost of estimating probability distributions for large datasets, but modern advancements, such as variational inference and Monte Carlo methods, have greatly improved their scalability \cite{wu2018deterministic}. Furthermore, the rise of deep learning has brought about new probabilistic architectures, like Bayesian neural networks, which build on the foundational ideas of PNNs. As research continues, integrating probabilistic reasoning into neural network models is expected to play an even larger role in fields like autonomous systems, where safety and uncertainty are critical considerations \cite{feng2021review}.

Probabilistic neural networks have numerous applications across a wide range of fields. In finance, for example, they are used for risk assessment \cite{anbari2017risk} and modelling uncertain market trends \cite{faes2020recent}. In healthcare, PNNs assist in medical diagnostics, where the uncertainty in patient data makes probabilistic models particularly valuable \cite{varuna2018identification}. They are also employed in robotics \cite{sun2017indoor} for decision-making under uncertainty \cite{wozniak2018small} and in natural language processing for tasks like sentiment analysis \cite{ain2017sentiment}, where multiple interpretations of text are possible. PNNs' ability to manage uncertainty makes them ideal for tasks where traditional deterministic models might struggle due to ambiguous or incomplete data. PNNs are widely used in other domains, such as interval data classification \cite{kowalski2017interval} and stream data classification \cite{rutkowska2024probabilistic}, where the nature of the input data requires flexible probabilistic handling. Due to their structure and explainable nature, PNNs enable controlled dimensionality reduction \cite{Kowalski_determ} and dataset size pruning \cite{kowalski2017sensitivity}, allowing them to efficiently handle high-dimensional data while preserving the interpretability of the model. This controlled reduction makes PNNs a powerful tool in data-intensive applications where it is crucial to balance model complexity with performance, especially when working with large or streaming datasets.

\subsection{Structure of probabilistic neural network}
The functioning of the PNN classifier is built upon the kernel density estimator (KDE), which is a non-parametric method for estimating the probability density function of a given dataset. In general, the KDE can be expressed as:

\begin{equation}
\hat{f}(\mathbf{x})=\frac{1}{Ph^N}\sum_{p=1}^{P}{K\left( \frac{\mathbf{x}-\mathbf{x}^{(p)}}{h} \right)}.
\label{eq:pnn_KDE}
\end{equation}

Here,  $\mathbf{x}=[x_{1},\ldots, x_{N}]$ represents a test sample, $h$  is the smoothing parameter, and $K\left(\cdot \right)$ is the kernel function that maps input data from 
$\mathbb{R}^{N}\to [0,\infty)$. The kernel function is responsible for quantifying the similarity between the test point and each training point in the dataset. The parameter $h$
controls the width of the kernel and, therefore, how much influence each training point has on the estimate.
For multi-dimensional datasets, the kernel density estimator is often generalised using a product of individual kernels applied to each dimension of the input data:

\begin{equation}
K(\mathbf{x})=\mathcal{K}(x_1)\cdotp \mathcal{K}(x_2)\cdotp \dots\cdotp\mathcal{K}(x_N).
\label{eq:produkt}
\end{equation}
In the 
%presented 
approach presented in this paper, the kernel function for each dimension, $\mathcal{K}(x_i)$, is defined as:

\begin{equation}
\mathcal{K}(x_i)=\frac{2}{\pi(x_i^2+1)^2} .
\label{eq:Cauchy}
\end{equation}

This specific form of the kernel is a one-dimensional Cauchy function chosen because of its beneficial analytic properties, particularly in terms of ensuring a well-behaved derivative. This kernel function is useful when it comes to modelling data that has heavier tails, as it places less emphasis on distant data points compared to other kernels like the Gaussian one.

It is also important to note that the exact choice of the kernel function can have a minor impact on the accuracy of the density estimation, with research suggesting that the kernel selection typically causes around a 4\% difference in the quality of the non-parametric estimate. However, the kernel function choice should be tailored to the specific requirements of the application at hand. In this particular implementation of PNN, the product form of KDE (\ref{eq:produkt}) along with the Cauchy kernel (\ref{eq:Cauchy}) is used to model the underlying data distribution. This configuration was selected for its analytical convenience and suitability for the 
%given 
considered classification task.

A PNN is structured with four distinct layers that work together to classify or predict outputs based on the input data. The first layer is the \textit{input layer}, where the attributes of the input vector \(x\) are fed into the network. This layer simply passes the input data to the next layer without any transformation or computation.

The second layer is the \textit{pattern layer}, which contains one neuron for each training sample in the dataset. In this layer, each neuron represents a training example and computes a similarity measure between the input vector and that specific example. The neurons in this layer compare the input vector to each training point, generating an output that reflects how closely the input matches each training example. There are two common approaches to computing these similarities: radial basis functions (or radial kernels) and product kernels. In the proposed case, the product kernel approach is used to calculate the output of the pattern layer.

The third layer is the \textit{summation layer}, which aggregates the outputs from the pattern layer. In this layer, there is one neuron for each class \(j\), and each of these neurons collects signals from the pattern neurons that correspond to training examples of the 
%same 
respective class. The summation neuron for a given class computes the probability density estimate for the input vector belonging to that class (\ref{eq:pnn_KDE_j}). This is done using a formula that integrates several key parameters, such as the number of training examples \(P_j\) in the \(j\)-th class, a matrix of smoothing parameters \(h\), and a modification coefficient \(s_p\). The output of each summation neuron reflects the likelihood that the input vector belongs to its respective class.

\begin{equation}
\hat{f}_j(\mathbf{x})=\frac{1}{P_j{\text{det}(\mathbf{h})}}\sum_{p=1}^{P_j}{\frac{1}{s_p^N}K\left({\frac{\left( \mathbf{x}-\mathbf{x}^{(p)}_j\right) ^T\mathbf{h}^{-1}}{s_p}}\right)},
\label{eq:pnn_KDE_j}
\end{equation}

The \textit{output layer} consists of a single neuron that makes the final decision about the input's class. The neuron selects the class with the highest output from the summation layer, effectively applying Bayes' theorem to determine the most probable classification for the input vector. The decision is made by selecting the class \(Out(x)\) that maximises the estimated probability density function \(\hat{f}_j(x)\) across all classes \(j\) (\ref{eq:pnn_output_layer}).

\begin{equation}
Out(\mathbf{x}) = \underset{j=1\ldots J} {\mathrm{argmax}} ~\hat{f}_j(\mathbf{x}),
\label{eq:pnn_output_layer} 
\end{equation}

The training process for a PNN primarily revolves around selecting appropriate values for the smoothing parameters \(h_i\) and the modification coefficients \(s_p\), which control the behavior of the pattern neurons and summation neurons, respectively. These parameters are critical for ensuring that the PNN accurately models the underlying probability distributions of the training data.

\subsection{Training procedures }

The process of training 
%Probabilistic Neural Networks
PNNs, particularly in determining the smoothing parameters, encompasses a variety of methodologies that can be broadly categorised into two distinct groups. The first group consists of statistical approaches, such as cross-validation and plug-in methods. These techniques stem from classical statistics and aim to optimise the smoothing parameter by minimising the mean squared error between the nonparametric probability density estimation and the true probability density function. This minimisation ensures that the density estimation closely aligns with the underlying data distribution, making these methods particularly effective in scenarios where statistical accuracy of density estimation is the primary objective.

However, when PNNs are employed in classification tasks, the focus shifts from pure statistical accuracy to achieving optimal class separability. In such cases, the primary goal is not merely to approximate the true probability density function but to position the density estimation functions of different classes in a way that maximises their distinction. This ensures that the PNN classifier can effectively differentiate between classes, which is often more critical than minimizing 
%achieving minimal 
the error 
%in 
of density estimation.

To address this classification-specific objective, alternative approaches to training PNNs have been developed. These include methods based on metaheuristic optimisation algorithms, such as 
%genetic algorithms, particle swarm optimisation, 
GA, PSO or 
%simulated annealing
SA, which search for the optimal smoothing parameters by directly optimising classification performance metrics. Additionally, reinforcement learning techniques can be employed, where the training process iteratively adjusts the smoothing parameters based on the feedback from the classification task itself, ultimately learning the parameter values that yield the highest classification accuracy.

By tailoring the training process to the specific requirements of the classification task, these alternative approaches can overcome the limitations of purely statistical methods, enabling PNNs to achieve superior performance in complex classification scenarios.

A crucial element in training 
%Probabilistic Neural Networks 
PNNs is selecting an appropriate method for determining the smoothing parameter, as this choice profoundly affects the network's classification performance. The smoothing parameter plays a pivotal role in controlling the balance between overfitting and underfitting, making its adjustment critical to the success of the model. Various approaches to defining the smoothing parameter have been proposed, each offering different levels of flexibility and application scenarios.

The simplest approach involves using a single scalar value for the smoothing parameter $(h_I=h)$, applied uniformly to all pattern neurons in the network. This method is computationally efficient and is often preferred when the data is relatively homogeneous, and the classes are well-separated. However, its limitations become evident in more complex datasets, where a single global value may not adequately capture variations in data distribution. 

An alternative approach assigns a single smoothing parameter value to each class. This method introduces more flexibility by allowing the parameter to vary between classes, enabling better adaptation to the characteristics of each class distribution. This approach is particularly useful in datasets where the classes exhibit distinct densities or variances, as it ensures that each class is smoothed appropriately without being overly rigid or 
%excessively 
too complex. This level of granularity can be expressed using the following formula:

\begin{equation}
\mathbf{h_{II}} = [h^{(1)}, h^{(2)}, \dots, h^{(G)}]
\label{h_II}
\end{equation}
where \( h^{(g)} \) represents the smoothing parameter for the \( g \)-th class, and \( G \) is the total number of classes in the dataset.

For even greater flexibility, a vector-based smoothing parameter can be employed, where each coordinate of the input pattern has its own smoothing value. This method is advantageous in high-dimensional datasets where different features exhibit varying levels of relevance or variability. By tailoring the smoothing parameter to each feature, the network can better adapt to local data structures and improve classification performance in such settings. This approach can be expressed using the following formula:

\begin{equation}
\mathbf{h_{III}} = [h_1, h_2, \dots, h_n],
\label{h_III}
\end{equation}
where \( h_j \) represents the smoothing parameter for the \( j \)-th coordinate of the input pattern, and \( n \) is the total number of features in the dataset.

Finally, the most advanced approach involves a matrix of smoothing parameters, where each coordinate has a unique value not only for the feature but also for the class it belongs to. This approach provides the highest level of customisation, allowing the network to account for intricate interdependencies between features and class distributions. It is particularly beneficial in complex classification tasks where class-specific feature relationships are essential for accurate predictions. This method can be expressed using the following formula:

\begin{equation}
\mathbf{H_{IV}} = 
\begin{bmatrix}
h^{(1)}_1 & h^{(1)}_2 & \dots & h^{(1)}_n \\
h^{(2)}_1 & h^{(2)}_2 & \dots & h^{(2)}_n \\
\vdots & \vdots & \ddots & \vdots \\
h^{(G)}_1 & h^{(G)}_2 & \dots & h^{(G)}_n
\end{bmatrix}
\label{h_IV}
\end{equation}
where \( h^{(g)}_j \) represents the smoothing parameter for the \( j \)-th coordinate of the input pattern and the \( g \)-th class, and \( G \) is the total number of classes in the dataset. This matrix structure enables the network to apply a distinct smoothing parameter for each feature within each class, offering the highest degree of flexibility and precision in adapting to the data's complex structure.

The choice of smoothing parameter method depends on the nature of the dataset and the complexity of the classification task. Simpler approaches, such as the scalar or class-level parameters, are suitable for relatively uniform datasets with clear class separability, while more complex methods, such as vector or matrix parameters, are better suited for high-dimensional or heterogeneous datasets with overlapping or intricately distributed classes. Each method represents a trade-off between computational efficiency and the capacity to model complex data structures effectively.

The performance of each method for determining the smoothing parameter, as well as the chosen configuration for the number of smoothing parameters, largely depends on the specific task assigned to the PNN. 
%A detailed review of t
The effectiveness of various smoothing parameter selection procedures has been thoroughly analysed in the following articles \cite{kowalski2020probabilistic, kowalski2023feature}.

\subsubsection*{Smoothing parameter modification procedure}

Once the temporary values of the smoothing parameter vector are obtained using one of the methods presented above,
%methods, 
the KDE quantities are calculated based on 
%equation 
(\ref{eq:pnn_KDE_j}) for each element \( x^{(p)} \) where \( p = 1, \dots, P \). This enables 
%us 
to independently compute the modification parameter \( s_p \) for all \( x^{(p)} \) patterns using the formula:

\begin{equation}
s_p = \left( \frac{\hat{f}(x^{(p)})}{ \tilde{s}} \right)^{-c},
\end{equation}
where \( \tilde{s} \) represents the geometric mean of the KDE values \( \hat{f}(x^{(p)}) \), and \( c \geq 0 \) is a constant determining the intensity of the modification. As \( c \) increases, the modification intensity grows. It is important to note that when \( c = 0 \), \( s_p \equiv 1 \), meaning no modification is applied to the smoothing parameter.
The primary goal of introducing the smoothing parameter modification procedure in %Probabilistic Neural Networks 
PNNs is to adjust the level of smoothing for individual data points to enhance the quality of density estimation. This procedure enables dynamic modification of the smoothing parameter based on the local properties of the data, such as the density of observations. It allows for more precise modelling of diverse data structures while minimising the effects of over-smoothing or under-smoothing.

\section{The \textit{constrained Hybrid Metaheuristic} (cHM) algorithm}
\label{sec:cHM}

\subsection{Proposed cHM algorithm}
%Here, w
We propose a constrained Hybrid Metaheuristic (cHM) %procedure, 
algorithm that combines 
%multiple 
several swarm-based optimisation algorithms into a 
%super-metaheuristic 
coherent metaheuristic method. For the sake of clarity of the presentation, we 
%his work 
will 
%also 
refer to the component swarm-based procedures (operating %inside the 
withinn cHM) as \textit{weak metaheuristics}, \textit{inside-optimisers} or \textit{single metaheuristics}. 
%The presented technique can 
cHM uses several weak metaheuristics 
%algorithms 
based on a population of individuals, in particular, the ones mentioned in the previous section.
%e.g., PSO and FPA. 
In 
%this research, 
each of them, a population is a group of individuals that represent 
%a set of 
potential solutions, i.e. 
%possible 
smoothing parameter vectors for a given PNN. Each individual 
%is 
contains sufficient information to produce a functional PNN. 
%The population is built to be shared between all algorithms that are part of the cHM. 

The proposed optimisation method consists of two phases that can be repeated \textit{n}-times: \textit{probing} and \textit{fit}. These phases are constrained in execution by the maximum number of times each phase calls an evaluation (fitness) function, \textit{$maxFE_{probing}$} and \textit{$maxFE_{fit}$} respectively. The $maxFE{probing/fit}$~\cite{cec} constraint could be transformed into other limitations, for instance, a time-based evaluation, which, however, strongly depends on the
%In fact, for this study, the value of the training phase $maxFE{probing/fit}$ was considered the most reliable constraint type as it is not affected by the 
computational resources used in the experiments. It should be mentioned that, in this research, $maxFE_{probing/fit}$ 
%in this research calculates 
counts a single evaluation of each test sample of each individual as a separate evaluation. For instance, when the population of 10 individuals is evaluated with 100 samples, 1000 evaluations are added to the value $maxFE_{probing/fit}$.

In the first phase, the population for each weak metaheuristic is initialised similarly or taken from the previous iteration of the cHM algorithm. Next, each optimisation method is used separately to train the 
%Probabilistic Neural Network 
PNN until \textit{$maxFE_{probing}$} number of evaluations is not met. Then, the 
%single 
method with the lowest cost function value (\textit{error rate}) is selected for further PNN training. The population of the best single metaheuristic is saved, 
%and it will 
to be passed to the next phase. In the case of the same function cost scores 
%for 
tied by multiple 
%optimisation techniques
metaheuristics, the best one of them is selected randomly.
%from the best ones.

The second phase 
%takes 
considers the best-performing
%, single 
metaheuristic from the first phase. It uses the optimisation procedure, together with its population, to train 
%the probabilistic neural network 
PNN for the \textit{$maxFE_{fit}$} number of evaluations. In the end, after the PNN is finished, the metaheuristic population from this step is saved, 
%and it will 
to be passed to the next iteration of the 
%constrained Hybrid Metaheuristic 
cHM algorithm.

These two phases are repeated \textit{n}-times or until the 
%condition criteria are met. To exemplify, 
process converges, i.e., the \textit{error rate} 
%would be 
is equal to $0$ 
%for 
on the test set. 

The detailed cHM pseudocode 
%that describes the cHM 
is shown in Algorithm~\ref{algorithm:cHM}.

\begin{algorithm}
	\caption{constrained Hybrid Metaheuristic Optimization} 
   
	\begin{algorithmic}[1]
    \State Assume $k$ metaheuristics with common characteristics $\theta_n$, that describe a population of solutions for a Probabilistic Neural Network (PNN).
    PNN will be trained in a constrained number of maximum Function Evaluations ($maxFE_{probing}$, $maxFE_{fit}$) to ensure reliable metaheuristics performance comparison.
    \For {$n=1,2,\ldots n$}
        \State Begin with metaheuristics probing
        \For {$k=1,2,\ldots k$}
            \State Initialize $maxFE_{1}$ = 0.
            \While {${maxFE_{1}} < maxFE_{probing}$}
                \State Train PNN with each $k-th$ metaheuristic
                \State Increment $maxFE_{1}$ according to the number of function evaluations for probing with $k-th$ method
            \EndWhile           
        \EndFor
        \State Select the best-performing $k-th$ metaheuristic and update $\theta_n$
        \State Initialize $maxFE_{2}$ = 0.
        \While {${maxFE_{2}} < maxFE_{fit}$}
            \State Train PNN with $k-th$ metaheuristic using $\theta_n$ parameters 
            \State Increment $maxFE_{2}$ according to the number of function evaluations during fit phase 
        \EndWhile
        \State Update $\theta_n$ 
        \If{PNN convergence is met}
            \State Break
        \EndIf
        \State Reset metaheuristics parameters and pass the best population to the next probing phase
	\EndFor
    \\
    {Return PNN smoothing parameters an individual with the best metric value from $\theta_{best}$}
	\end{algorithmic} 
    \label{algorithm:cHM}
\end{algorithm}

\subsection{Metaheuristic procedures}
%Recently, nature-based computational intelligence (CI) algorithms have played a vital role in optimising different problems. From constrained optimisation, to training artificial neural networks, and optimising technical processes \cite{comprehensive_pso}. Indeed, training PNN was proven to be especially effective with some metaheuristics, because of common characteristics of non-parametric behavior for both the PNN and the CI procedure \cite{BAT-laplacian} \cite{fpa_pnn_kowalski}.

Generally, when using swarm-based algorithms to train PNNs, an individual in a population has the form of a vector of proposed smoothing parameters, i.e., each individual ($h_{III}$) is a vector of parameters sufficient to trigger a PNN for a given data \cite{fpa_pnn_kowalski}. 

In the experiments with the proposed cHM algorithm the following portfolio of five metaheuristic methods have been considered: PSO, BAT algorithm (BAT), Bacterial Foraging Optimization (BFO), Simmulated Annealing (SA), and Flower Pollination Algorithm (FPA). All of these global optimization methods are well-known in the literature and have been described in numerous papers. In our implementation, the vanilla formulations of these metaheuristics are considered, and therefore, for the sake of space-saving, in what follows we only briefly mention the underlying principles and search mechanisms of these methods along with the relevant literature.

\textbf{PSO} is one of the most widely used nature-inspired algorithms, with multiple enhancements presented in the field. Original PSO formulation refers to a swarm-based technique founded on the cooperation of particles in a swarm (population) \cite{pso_origin}. The particles move around in a search space \textit{S} iteratively, looking for the optimal position. 
%Indeed, e
Each particle has its position \textit{p} and velocity \textit{v} that are updated through algorithm iterations. The new particle's position is influenced by its historically best position, as well as the historically best position of the entire swarm or the selected part of a swarm called the particles's neighborhood.

\textbf{BAT}, similarly to PSO,  procedure uses a population of individuals to 
%look through a 
search the space for a sub-optimal problem solution. It is inspired by bats' behavior for communication when hunting or moving. 
%They 
Bats use echolocation with varying frequency and loudness, depending on the distance from the prey and the size of the award~\cite{bat-core}. The BAT algorithm (BA) might be seen as a special case of PSO and has been used to train PNNs before~\cite{BAT-laplacian}.

\textbf{BFO} is based on the behavior of E. Coli bacteria foraging motions, and
%The Bacterial Foraging Optimization (BFO) 
models different movements of the E. Coli including chemotaxis, swarming, reproduction, elimination, and dispersal~\cite{bfo}. These procedures are responsible for the bacterium actuation, sensing, and decision-making processes. Similarly to PSO and BA, the BFO procedure 
%has 
maintains a population of individuals that iteratively seek an optimal solution to the problem in a given space.

\textbf{SA} is inspired by the annealing phenomenon while crystals are grown from melt~\cite{sa}. In SA, similarly to previously-mentioned techniques, a population of individuals is used to search a solution space. The particles are initialized randomly, and the algorithm flow is controlled by two factors: temperature $T$ and the Boltzmann distribution. Over the SA iterations, a new potential particle position is accepted 
%by the population 
if 
%their 
its cost function value is lower than previously (before the potential movement). Otherwise, the newly generated solution is accepted with the so-called acceptance probability, defined by the Boltzmann equation. This probability depends on the temperature, which gradually decreases in time, thus cooling down the SA process.

\textbf{FPA} is inspired by insect flower pollination. 
%this algorithm 
The method iteratively searches the space of possible solutions by combining the following two steps: local (exploration) and global (exploitation) optimisation~\cite{fpa}. 
%and is based on two main parts: local (exploration) and global (exploitation) optimisation. 
%FPA iteratively searches over the space of possible solutions by combining these two steps \cite{fpa}. 
A random variable $r$  and parameter $p$ control the procedure's flow. For each individual, if the $r > p$ the exploration phase is turned on. Otherwise,
%\cite{fpa}. It uses the Levy distribution to mimic insects flying over long distances \cite{fpa}.
%
%On the other hand, when $r$ is lower than $p$, 
a local examination is performed randomly, around the current position of the individual. This exploitation step is referred to as ``self-pollination''~\cite{fpa_pnn_kowalski}

\section{Experimental results}
\label{sec:Results}

The cHM method 
%is assumed to merge 
combines several metaheuristics into one optimisation method and 
%In fact, the hybrid method is supposed to 
leverages the advantages of each particular swarm-based technique in a shorter time frame. In addition, 
%its 
cHM application may
%might 
lead to low-cost evaluation of different methods for a given problem. 
%To test its performance, 
In the experiments, 
%the 
cHM was used for training PNN,
%Probabilistic Neural Network training. 
with BAT, SA, FPA, BFO and PSO procedures 
%were 
%put into the main training algorithm 
used as inside optimisers in Algorithm~\ref{algorithm:cHM}. The methods used for the algorithm evaluation are presented in the following, together with the results of the experiment.

\subsection{PNN training details}
In this research, 
%the 
PNNs were constructed using the Cauchy kernel with separate smoothing parameters for each feature vector \(h_{III}\) (\ref{h_III}) in the dataset. The 
%constrained Hybrid Metaheuristic (
cHM
%) technique 
algorithm was employed to train the PNNs for classification problems. 
%The 
cHM 
%algorithm 
was initialized with a randomly generated population of individuals, each representing a set of smoothing parameters required to build the 
%neural network
PNN. The initial population consisted of 20 individuals with values constrained to real numbers in the range \([0, 10]\). This same population was used to initialize each of the %weak metaheuristics 
inside optimizers within the cHM method. To ensure the reproducibility of experiments, a fixed random seed was applied to all stochastic operations performed during the calculations.

When training PNNs with swarm intelligence methods, the smoothing parameters are determined using heuristic
%ally defined algorithms
methods. 
%To account for the limitations of these neural networks, t
The possible smoothing parameters were constrained to the interval \([0, 10000]\) of real numbers. If the \(h_i\) value was negative, the reflection technique \cite{reflection} was applied to ensure the value fell within the constrained range of positive numbers.

The parameters of the cHM algorithm are shown in Table \ref{table:chm_params}. The parameters used for each metaheuristic inside the cHM are shown in Tables \ref{table:BAT_params} - \ref{table:SA_params}. These parameters were selected 
%on the basis of 
according to the referenced papers.
\begin{table}[!h]
\centering
\centering
\begin{tabular}{|c|c|c|c|c|c|}
\hline
$metaheuristics$                                                   & $n$ & $n_p$ & $f. threshold$ & $maxFE_{probing}$     & $maxFE_{fit}$      \\ \hline
\begin{tabular}[c]{@{}c@{}}PSO, FPA, BAT, \\ BFO, SA\end{tabular} & 5          & 20            & 1e-8                  & $n_p$ * $n_{t}$ * 30 & $n_p$ * $n_{t}$ * 100 \\ \hline
\end{tabular}
\caption{Parameters of the cHM algorithm. 
%The 
$n$ is the number of cHM iterations, $n_p$ represents the number of individuals in a population, $n_t$ is the cardinality of the training sample, and $f.$ stands for fitness.}
\label{table:chm_params}
\end{table}
\begin{table}[!h]
\centering
\begin{tabular}{|c|c|c|c|c|}
\hline
$loudness$ & $\alpha$ & $\gamma$ & $min_f$ & $max_f$ \\ \hline
10                & 0.9            & 0.9            & 0              & 1           \\  \hline 
\end{tabular}
\caption{Parameters of the BAT algorithm \cite{BAT-laplacian}.}
\label{table:BAT_params}
\end{table}
\begin{table}[!h]
\centering
\begin{tabular}{|c|c|c|c|c|c|c|c|c|}
\hline
$ed_s$ & $C_i$ & P$_{ed}$ & $N_c$ & $N_s$ & $d_a$ & $w_a$ & $h_r$ & $
w_r$ \\ \hline
2    & 0.2 & 0.25     & 4   & 4   & 0.1 & 0.2 & 0.1  & 10   \\ \hline
\end{tabular}
\caption{Parameters of the BFO algorithm \cite{bfo}.}
\label{table:BFO_params}
\end{table}
\begin{table}[!h]
\centering
\begin{tabular}{|c|}
\hline
\textit{switch probability} \\ \hline
0.8                         \\ \hline
\end{tabular}
\caption{Parameters of the FPA algorithm \cite{fpa}.}
\label{table:FPA_params}
\end{table}
\begin{table}[!h]
\centering
\begin{tabular}{|c|c|c|c|}
\hline
$\omega$ & $c_1$ & $c_2$ & \textit{adjust  $\omega$}      \\ \hline
1 & 0.5  & 1    & \textit{true} \\ \hline
\end{tabular}
\caption{Parameters of the PSO algorithm \cite{pso_origin}.}
\label{table:PSO_params}
\end{table}
\begin{table}[!h]
\centering
\begin{tabular}{|c|c|c|c|}
\hline
$T$   & $\alpha$ & $s_T$ & $d$    \\ \hline
100 & 0.9   & 1e-8 & 0.01 \\ \hline
\end{tabular}
\caption{Parameters of the SA algorithm \cite{sa}.}
\label{table:SA_params}
\end{table}
Generally, to train 
%Neural Networks 
PNNs with a swarm-based method, a cost function is needed. Here, we used an error rate function,
%for this purpose, 
defined as follows:
\begin{equation}\label{equation:error_rate}
    \textit{error rate} = 1 - \frac{\textit{number of correct predictions}}{\textit{cardinality of test sample}}.
\end{equation}
The performance of PNN training with the cHM algorithm was tested on multiple datasets. Before training, each dataset was split into train and test sets in a stratified manner. The particular sets were then used to calculate train and test metrics to evaluate the methods on 
%unleaked 
unseen data. The test size was set to 20 $\%$ of the original data size. The training procedure for each dataset was repeated 10 times to calculate the cumulative (average) metrics: accuracy, precision, and recall \cite{scikit-learn}.

\subsection{Datasets}

Table~\ref{table:datasets} lists 
%the 
16 datasets used for the evaluation of the cHM algorithm for the PNN training. 
%The Table contains 16 dataset names, and different characteristics: cardinality, number of classes, number of features, and class distribution. 
The datasets come from the UCI ML repository \cite{uci}, kaggle \cite{ghouls-goblins-and-ghosts-boo} and the PMLB repository \cite{pmlb}. The datasets represent different variations of the classification problems. For example, there are datasets for binary (Cancer, Parkinson, Climate) and multiclass (Glass, Heart, Vehicle) classifications. In addition, the classes in various datasets have fairly balanced (Ghost, Banknote, Vecivle) or imbalanced (Parkinson, Climate) distributions. To exploit the feature level of the smoothing parameter in PNNs, the datasets have varying numbers of features (from 4 to 30). It is assumed that this set of characteristics helps with a comprehensive comparison of the cHM 
%technique 
algorithm application 
%for Probabilistic Neural Network 
to PNN training.
\begin{table}[!h]
\scalebox{0.75}
{
\begin{tabular}{|c|c|c|c|c|}
\hline
Dataset                           & \textbf{No. of rows} & \textbf{No. of features} & \textbf{No. of classes} & \textbf{Class balance}                  \\ \hline
\textbf{Iris}                     & 150                 & 4                        & 3                       & 50/50/50                                \\ \hline
\textbf{Ghost}                    & 371                 & 5                        & 3                       & 129/125/117                             \\ \hline
\textbf{Cancer}                   & 569                 & 30                       & 2                       & 357/212                                 \\ \hline
\textbf{Wine}                     & 178                 & 13                       & 3                       & 71/59/48                                \\ \hline
\textbf{ILPD}                     & 579                 & 10                       & 2                       & 414/165                                 \\ \hline
\textbf{Glass}                    & 214                 & 9                        & 6                       & 76/70/29/17/13/9                        \\ \hline
\textbf{Parkinson}                & 195                 & 22                       & 2                       & 147/48                                  \\ \hline
\textbf{E. coli}                  & 332                 & 7                        & 6                       & 143/77/52/35/20/                        \\ \hline
\textbf{Banknote}                 & 1372                & 4                        & 2                       & 762/610                                 \\ \hline
\textbf{Heart}                    & 303                 & 14                       & 5                       & 164/55/36/35/13                         \\ \hline
\textbf{Climate} & 540                 & 21                       & 2                       & 494/46                                  \\ \hline
\textbf{Blood Transfusion}       & 748                 & 5                        & 2                       & 570/178                                 \\ \hline
\textbf{Thyroid}                  & 215                 & 6                        & 3                       & 150/35/30                               \\ \hline
\textbf{Monks}                    & 415                 & 7                        & 2                       & 229/186                                 \\ \hline
\textbf{Vehicle}           & 846                 & 19                       & 4                       & 218/217/212/199                         \\ \hline
\textbf{Pima}                     & 768                 & 9                        & 2                       & 500/268                                 \\ \hline

\end{tabular}
}
\caption{
%Information about 
Characteristics of the 16 datasets used for the cHM algorithm evaluation.}
\label{table:datasets}
\end{table}

\subsection{Results of proposed learning procedure}
%
%This section describes the results of computational experiments that applied the cHM procedure to PNN training. The constrained Hybrid Metaheuristic method 
cHM was 
%used 
applied to train and test PNNs for 16 datasets from Table~\ref{table:datasets}. 
Each training procedure 
%had the 
was performed using 
%the same 
a common set of constraints and parameters, presented in Table~\ref{table:chm_params}.
Then, each of the weak swarm optimisation techniques from the cHM was used separately for the PNN training, providing the single-method baselines.
%to compare the training methods.  
%Here, 
Due to paper length limitations, in what follows, we present only 
%partial 
the main results. The rest of them 
%might 
can be found in the appendix.
%due to paper length limitation.

\subsubsection{Metaheuristic training methods performance}
Tables~\ref{table:comparison_avg_test_accuracy} - \ref{table:comparison_avg_test_recall} show a comparison of the PNN classification performance when training with cHM and single metaheuristics.

Additionally, in Table \ref{table:comparison_max_test_accuracy}, the results of the classical deterministic method for training the PNN network, namely Plug-in, are provided for comparison. Due to its deterministic nature, this method was compared with the best results of the heuristic algorithms. The accuracy comparison showed that Plug-in outperformed the individual heuristic algorithms, but it was not able to surpass the proposed cHM algorithm. In the examined ranking, the Plug-in method was the best in 5 out of 16 datasets, while cHM was the best in 6 out of 16 datasets. Hyperparameter optimization was not performed for the heuristics methods, including the cHM, used to train PNN. In fact, Hyperparameter Optimization (HPO) performed for these methods has significantly improved their ability to train neural networks \cite{abd2021advanced}. As HPO goes beyond the scope of this research, it was not performed here. It should be noted that it might improve the performance of heuristic techniques in PNN training.

Table~\ref{table:comparison_avg_test_accuracy} presents the average test accuracy comparison. The $Rank$ is the sum of times  
a given method returned the highest scores. In cases where multiple methods share the same highest score, the $Rank$ value is increased by 1 for each of them. This rule applies to all reported metrics, including accuracy, precision, and recall. To exemplify, the 
BAT method in Table~\ref{table:comparison_avg_test_accuracy} has the $Rank$ value equal to 3 as it had the highest scores for the Parkinson, Blood Transfusion and Vehicle datasets. 

In Table~\ref{table:comparison_avg_test_accuracy}, it is shown that 
%the 
cHM 
%method gave 
yielded the highest value of the average test accuracy metrics more often than other methods. Indeed, 
%the Hybrid technique presented in this paper 
it outperformed other methods 
%for 
on both simple datasets like Iris, Breast Cancer \cite{uci}, or ILPD and more complex 
%data 
ones like Ecoli, Pima, and Climate model simulation datasets. cHM performed the worst on the Wine dataset, with the accuracy score of 0.789 compared to 0.878 for the BFO method. In addition, only the FPA method presented lower test metric value for this dataset than the cHM, 
which suggests that the population transmission between methods was not effective for this dataset. 

Tables~\ref{table:comparison_avg_test_precision} and~\ref{table:comparison_avg_test_recall} show the performance comparison of PNN training methods for the average test precision and the average recall metric, respectively. Similarly to 
%like for 
the accuracy metric, 
%the 
cHM overperformed other methods 
%by having 
and had the highest  $Rank$ values for these metrics. Consistently 
%and similar 
best results for the three different metrics show 
%that the cHM presented satisfactory 
the outstanding performance of the cHM algorithm 
%for different 
across various types of classification problems, especially those with class imbalance.

\begin{table}[!ht]
\centering
\scalebox{0.85}
{
\begin{tabular}{|c|c|c|c|c|c|c|c|}
\hline
\textbf{Dataset}                             & cHM & Plug-in & BAT & BFO & PSO & FPA & SA \\ \hline
\textbf{Iris}                                & 0.967        & 0.933            & 0.967        & 0.933        & 0.933        & 0.967        & 0.933       \\ \hline
\textbf{Banknote}  & 1.0          & 1.0              & 1.0          & 1.0          & 1.0          & 0.996        & 1.0         \\ \hline
\textbf{Ghost}                    & 0.653        & 0.613            & 0.64         & 0.613        & 0.653        & 0.56         & 0.613       \\ \hline
\textbf{Cancer}                              & 0.965        & 0.974            & 0.965        & 0.956        & 0.965        & 0.956        & 0.965       \\ \hline
\textbf{Wine}                                & 0.944        & 0.972            & 1.0          & 0.972        & 0.944        & 0.889        & 0.917       \\ \hline
\textbf{ILPD} & 0.672        & 0.647            & 0.664        & 0.69         & 0.69         & 0.681        & 0.681       \\ \hline
\textbf{Glass}                           & 0.674        & 0.721            & 0.698        & 0.581        & 0.605        & 0.674        & 0.651       \\ \hline
\textbf{Parkinson}                       & 0.949        & 0.974            & 0.949        & 0.949        & 0.897        & 0.949        & 0.949       \\ \hline
\textbf{E. coli}                           & 0.821        & 0.806            & 0.866        & 0.761        & 0.806        & 0.776        & 0.776       \\ \hline
\textbf{Heart}                           & 0.55         & 0.483            & 0.467        & 0.467        & 0.467        & 0.45         & 0.467       \\ \hline
\textbf{Climate}      & 0.88         & 0.861            & 0.88         & 0.861        & 0.87         & 0.861        & 0.889       \\ \hline
\textbf{Blood transfusion}              & 0.727        & 0.673            & 0.707        & 0.707        & 0.707        & 0.7          & 0.707       \\ \hline
\textbf{Thyroid}                         & 0.977        & 0.93             & 0.953        & 0.953        & 0.953        & 0.953        & 0.953       \\ \hline
\textbf{Monks}                           & 0.651        & 0.482            & 0.627        & 0.663        & 0.639        & 0.651        & 0.651       \\ \hline
\textbf{Vehicle}                         & 0.676        & 0.688            & 0.665        & 0.688        & 0.688        & 0.665        & 0.671       \\ \hline
\textbf{Pima}                            & 0.773        & 0.63             & 0.76         & 0.773        & 0.76         & 0.76         & 0.786       \\ \hline
\textit{Rank}                   & 6            & 5                & 4            & 4            & 4            & 1            & 3           \\ \hline
\end{tabular}
}
\caption{Comparison of PNN training methods for the max test accuracy metric.}
\label{table:comparison_max_test_accuracy}
\end{table}

\begin{table}[!ht]
\centering
\scalebox{0.95}
{
\begin{tabular}{|c|c|c|c|c|c|c|}
\hline
\textbf{Dataset}                         & cHM         & BAT        & BFO        & PSO        & FPA        & SA         \\ \hline
\textbf{Iris}                            & 0.927       & 0.897      & 0.927      & 0.917      & 0.92       & 0.927      \\ \hline
\textbf{Banknote}  & 0.993       & 0.986      & 0.97       & 0.993      & 0.949      & 0.963      \\ \hline
\textbf{Ghost}                    & 0.548       & 0.521      & 0.524      & 0.533      & 0.484      & 0.54       \\ \hline
\textbf{Cancer}                          & 0.954       & 0.947      & 0.946      & 0.952      & 0.939      & 0.945      \\ \hline
\textbf{Wine}                            & 0.789       & 0.831      & 0.878      & 0.8        & 0.767      & 0.844      \\ \hline
\textbf{ILPD} & 0.64        & 0.636      & 0.651      & 0.659      & 0.65       & 0.65       \\ \hline
\textbf{Glass}                           & 0.556       & 0.514      & 0.463      & 0.498      & 0.53       & 0.407      \\ \hline
\textbf{Parkinson}                       & 0.91        & 0.918      & 0.903      & 0.846      & 0.892      & 0.91       \\ \hline
\textbf{E. coli}                           & 0.734       & 0.722      & 0.631      & 0.627      & 0.67       & 0.672      \\ \hline
\textbf{Heart}                           & 0.437       & 0.398      & 0.39       & 0.398      & 0.398      & 0.397      \\ \hline
\textbf{Climate}      & 0.855       & 0.852      & 0.847      & 0.854      & 0.847      & 0.845      \\ \hline
\textbf{Blood transfusion}              & 0.689       & 0.695      & 0.692      & 0.691      & 0.688      & 0.693      \\ \hline
\textbf{Thyroid}                         & 0.953       & 0.923      & 0.926      & 0.93       & 0.928      & 0.921      \\ \hline
\textbf{Monks}                           & 0.577       & 0.564      & 0.62       & 0.554      & 0.576      & 0.599      \\ \hline
\textbf{Vehicle}                         & 0.641       & 0.652      & 0.636      & 0.642      & 0.651      & 0.626      \\ \hline
\textbf{Pima}                            & 0.748       & 0.716      & 0.732      & 0.717      & 0.718      & 0.723      \\ \hline
\textit{Rank}                        & \textit{10} & \textit{3} & \textit{3} & \textit{2} & \textit{0} & \textit{1} \\ \hline
\end{tabular}
}
\caption{Comparison of PNN training methods for the average test accuracy metric.}
\label{table:comparison_avg_test_accuracy}
\end{table}

\begin{table}[!h]
\centering
\scalebox{0.95}
{
\begin{tabular}{|c|c|c|c|c|c|c|}
\hline
\textbf{Dataset}                         & cHM        & BAT        & BFO        & PSO        & FPA        & SA         \\ \hline
\textbf{Iris}                            & 0.928      & 0.899      & 0.927      & 0.916      & 0.921      & 0.927      \\ \hline
\textbf{Banknote}  & 0.993      & 0.986      & 0.969      & 0.992      & 0.948      & 0.963      \\ \hline
\textbf{Ghost}                    & 0.565      & 0.524      & 0.544      & 0.537      & 0.498      & 0.551      \\ \hline
\textbf{Cancer}                          & 0.954      & 0.947      & 0.946      & 0.95       & 0.94       & 0.945      \\ \hline
\textbf{Wine}                            & 0.792      & 0.844      & 0.887      & 0.812      & 0.787      & 0.853      \\ \hline
\textbf{ILPD} & 0.57       & 0.576      & 0.588      & 0.594      & 0.585      & 0.585      \\ \hline
\textbf{Glass}                           & 0.536      & 0.519      & 0.355      & 0.435      & 0.494      & 0.332      \\ \hline
\textbf{Parkinson}                       & 0.878      & 0.887      & 0.865      & 0.798      & 0.854      & 0.874      \\ \hline
\textbf{E. coli}                           & 0.663      & 0.595      & 0.423      & 0.556      & 0.552      & 0.513      \\ \hline
\textbf{Heart}                           & 0.294      & 0.229      & 0.229      & 0.237      & 0.237      & 0.233      \\ \hline
\textbf{Climate}      & 0.471      & 0.49       & 0.505      & 0.497      & 0.505      & 0.488      \\ \hline
\textbf{Blood transfusion}              & 0.552      & 0.562      & 0.561      & 0.555      & 0.553      & 0.56       \\ \hline
\textbf{Thyroid}                         & 0.959      & 0.932      & 0.922      & 0.914      & 0.933      & 0.943      \\ \hline
\textbf{Monks}                           & 0.57       & 0.557      & 0.616      & 0.544      & 0.57       & 0.592      \\ \hline
\textbf{Vehicle}                         & 0.633      & 0.647      & 0.635      & 0.638      & 0.643      & 0.617      \\ \hline
\textbf{Pima}                            & 0.725      & 0.686      & 0.706      & 0.688      & 0.69       & 0.696      \\ \hline
\textit{Rank}                        & \textit{9} & \textit{3} & \textit{3} & \textit{1} & \textit{1} & \textit{0} \\ \hline
\end{tabular}
}
\caption{Comparison of PNN training methods for the average test precision metric.}
\label{table:comparison_avg_test_precision}
\end{table}

\begin{table}[!h]
\centering
\scalebox{0.95}
{
\begin{tabular}{|c|c|c|c|c|c|c|}
\hline
\textbf{Dataset}                         & cHM        & BAT        & BFO        & PSO        & FPA        & SA         \\ \hline
\textbf{Iris}                            & 0.927      & 0.897      & 0.927      & 0.917      & 0.92       & 0.927      \\ \hline
\textbf{Banknote}  & 0.994      & 0.987      & 0.97       & 0.993      & 0.949      & 0.963      \\ \hline
\textbf{Ghost}                    & 0.549      & 0.523      & 0.525      & 0.535      & 0.486      & 0.542      \\ \hline
\textbf{Cancer}                          & 0.948      & 0.939      & 0.939      & 0.946      & 0.929      & 0.936      \\ \hline
\textbf{Wine}                            & 0.788      & 0.831      & 0.881      & 0.802      & 0.771      & 0.847      \\ \hline
\textbf{ILPD} & 0.573      & 0.583      & 0.594      & 0.6        & 0.592      & 0.59       \\ \hline
\textbf{Glass}                           & 0.457      & 0.509      & 0.343      & 0.417      & 0.455      & 0.298      \\ \hline
\textbf{Parkinson}                       & 0.897      & 0.906      & 0.905      & 0.815      & 0.888      & 0.913      \\ \hline
\textbf{E. coli}                           & 0.504      & 0.503      & 0.37       & 0.381      & 0.442      & 0.371      \\ \hline
\textbf{Heart}                           & 0.251      & 0.211      & 0.208      & 0.214      & 0.208      & 0.206      \\ \hline
\textbf{Climate}      & 0.481      & 0.495      & 0.508      & 0.501      & 0.508      & 0.486      \\ \hline
\textbf{Blood transfusion}              & 0.546      & 0.555      & 0.555      & 0.548      & 0.548      & 0.553      \\ \hline
\textbf{Thyroid}                         & 0.916      & 0.845      & 0.865      & 0.89       & 0.865      & 0.835      \\ \hline
\textbf{Monks}                           & 0.568      & 0.556      & 0.613      & 0.544      & 0.57       & 0.589      \\ \hline
\textbf{Vehicle}                         & 0.642      & 0.653      & 0.638      & 0.644      & 0.653      & 0.628      \\ \hline
\textbf{Pima}                            & 0.708      & 0.669      & 0.686      & 0.677      & 0.674      & 0.675      \\ \hline
\textit{Rank}                        & \textit{8} & \textit{3} & \textit{5} & \textit{1} & \textit{2} & \textit{2} \\ \hline
\end{tabular}
}
\caption{Comparison of PNN training methods for the average test recall metric.}
\label{table:comparison_avg_test_recall}
\end{table}

\begin{figure}[H]
\centerline{\includegraphics[scale=0.6]{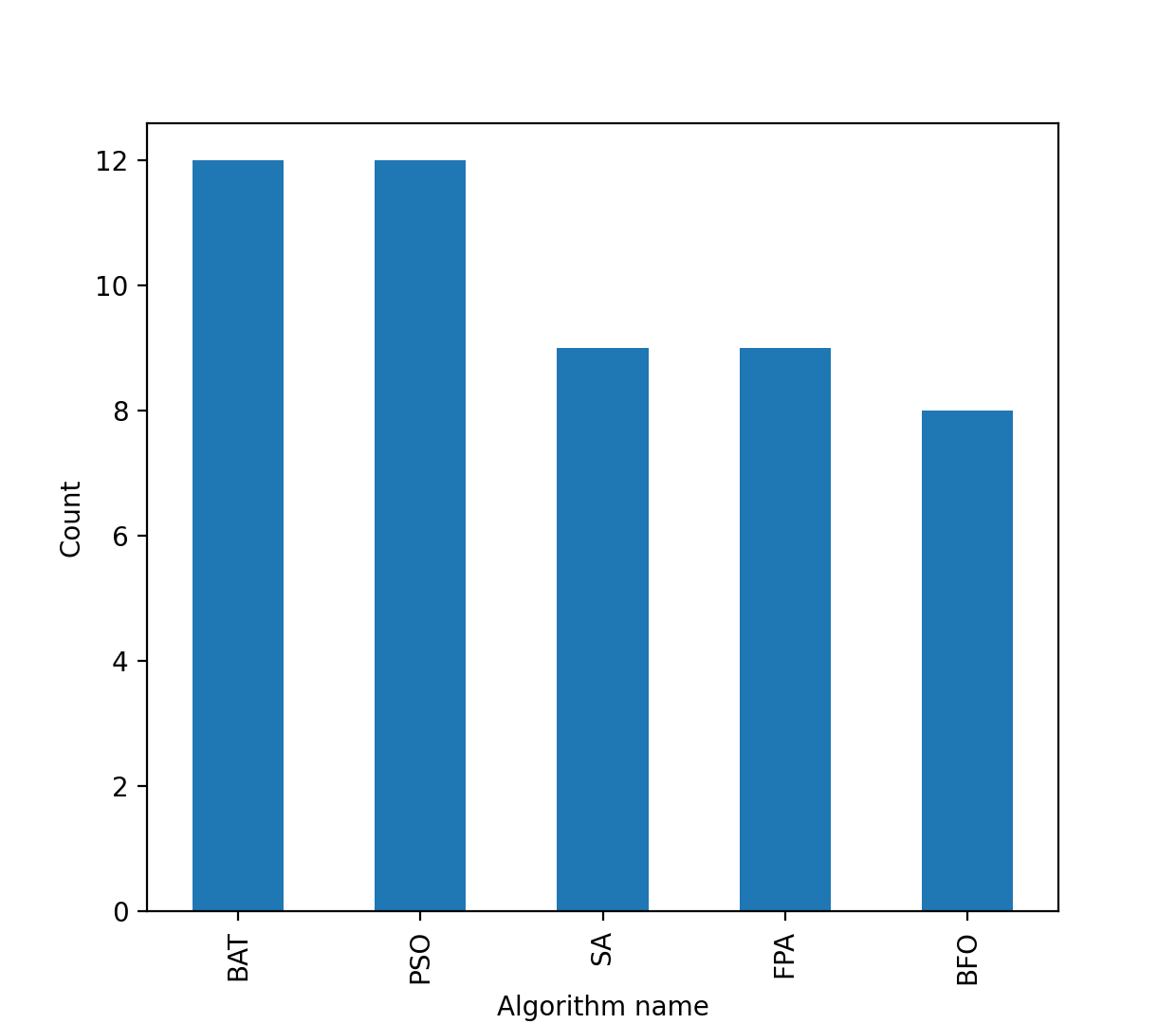}}
\caption{Bar plot of single metaheuristic selection by the cHM algorithm for the Cancer dataset.} 
\label{fig:bar_plot_cancer}
\end{figure}

\begin{figure}[H]
\centerline{\includegraphics[width=\linewidth]{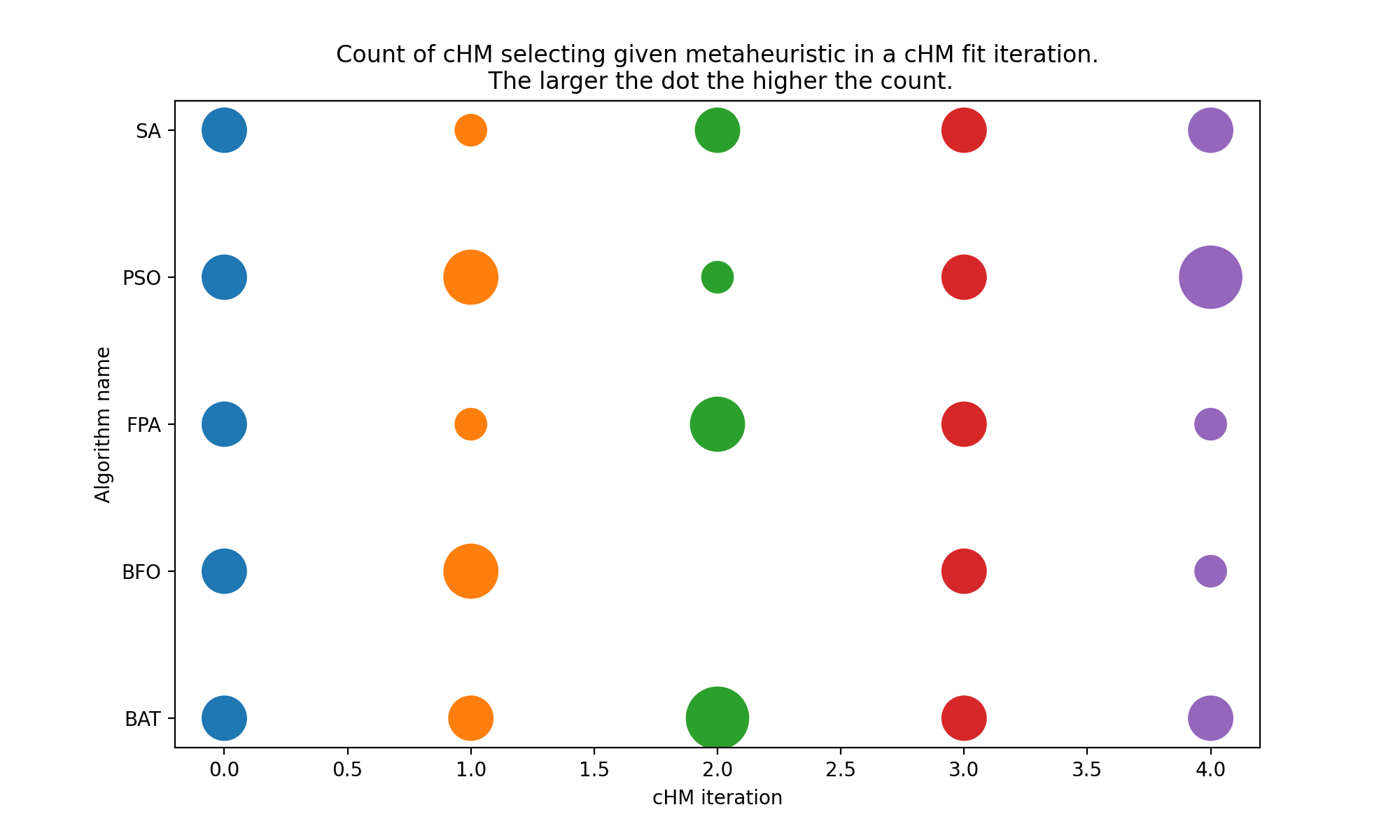}}
\caption{Dot plot of single metaheuristic selection by the cHM algorithm for the Cancer dataset. The count is presented for each iteration of the cHM method.}
\label{fig:dot_plot_cancer}
\end{figure}

\begin{figure}[H]
\centerline{\includegraphics[scale=0.6]{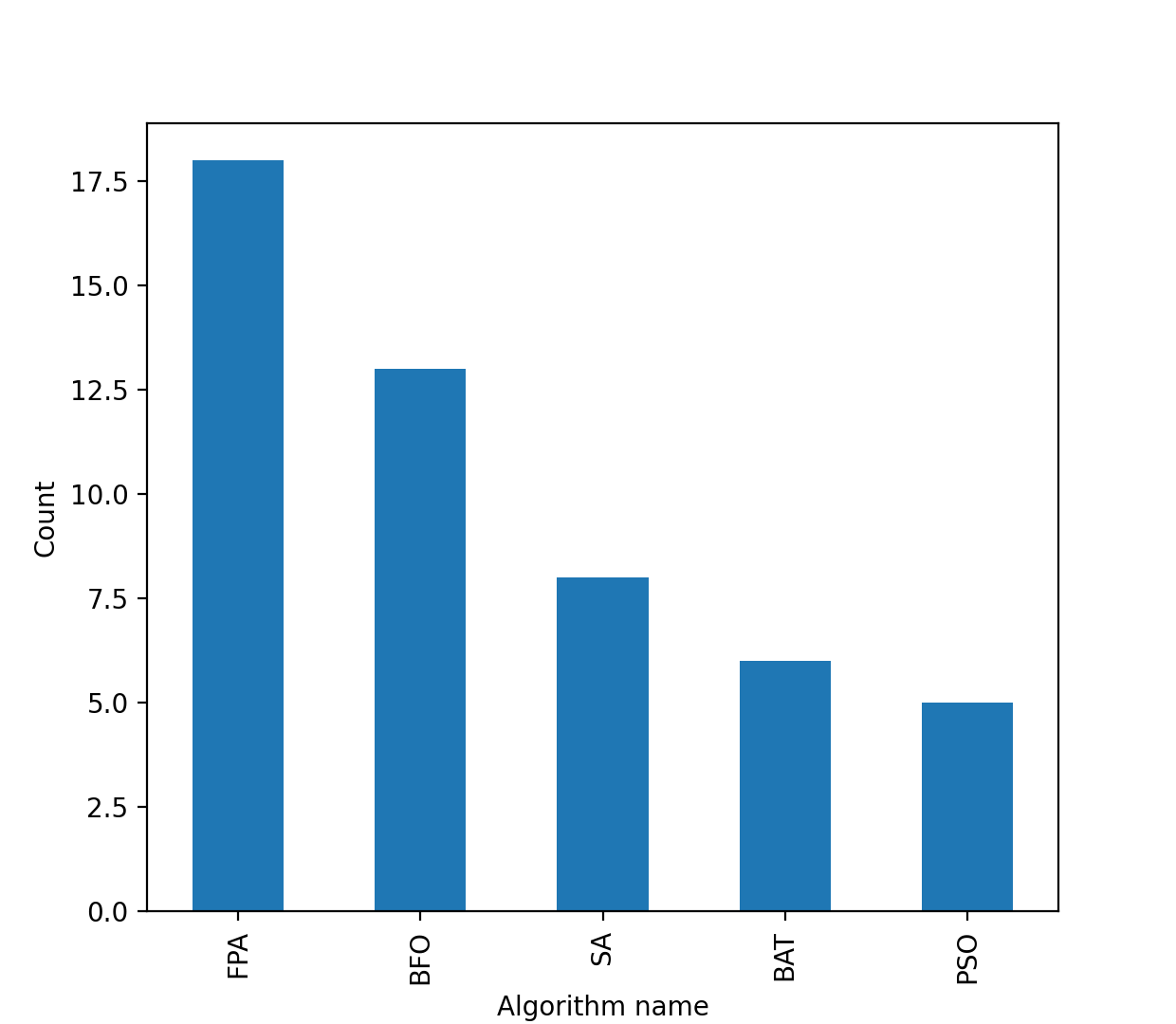}}
\caption{Bar plot of single metaheuristic selection by the cHM algorithm for the Vehicle dataset.} 
\label{fig:bar_plot_vehicle}
\end{figure}

\begin{figure}[H]
\centerline{\includegraphics[width=\linewidth]{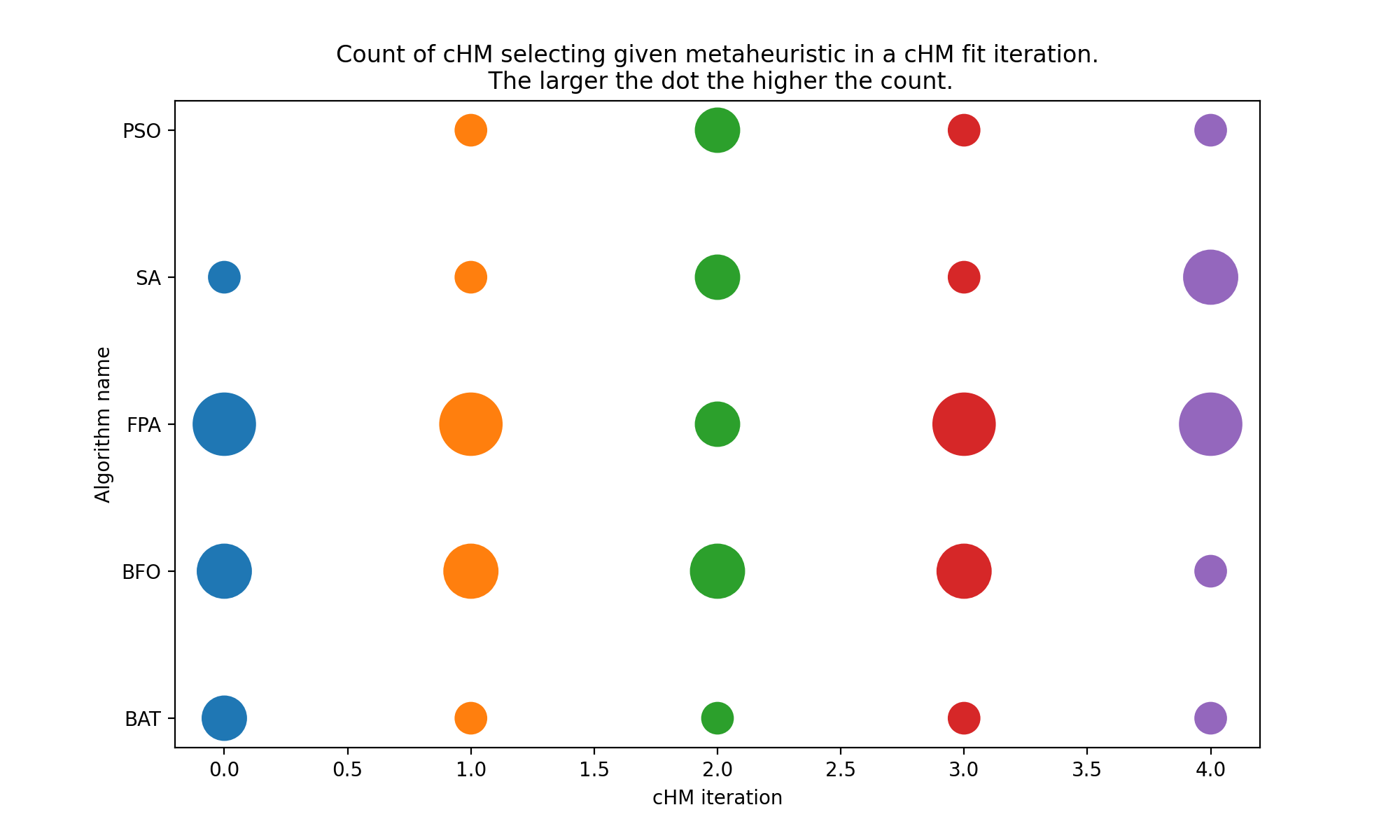}}
\caption{Dot plot of single metaheuristic selection by the cHM algorithm for the Vehicle dataset. The count is presented for each iteration of the cHM method.}
\label{fig:dot_plot_vehicle}
\end{figure}

\subsubsection{Metaheuristic selection frequency}
%\textbf{czy czestosc wybierania mety w cHM byla zgodna z tym ktora single dala najlepszy wynik}

Figure~\ref{fig:bar_plot_cancer} illustrates the frequency of a single metaheuristic selection for the cHM algorithm for the Cancer dataset. It can be noticed that although BAT and PSO were the most frequently chosen, there 
%is 
are no hard differences between method selections. 
%Indeed, 
All weak-optimisers of the 
%Hybrid technique 
cHM were selected during training.
%for this dataset. 
In addition, Figure~\ref{fig:dot_plot_cancer} shows the selection of a swarm method in the cHM algorithm for a specific iteration. As the experiment was repeated 10 times for each dataset, the size of the dot in 
%the 
Figure~\ref{fig:dot_plot_cancer} 
%is the measure of 
presents how often 
%the single meta 
a given metaheuristic was selected 
%by the constraint Hybrid Metaheuristic 
in a given iteration. Similarly, to 
%the histogram 
Figure~\ref{fig:bar_plot_cancer} the dot plot 
%illustrates 
demonstrates that 
%the 
cHM 
%method 
selects different inside optimisers evenly for the PNN classifier training of the Cancer dataset. It also shows, that the procedure 
%can use 
uses all weak methods 
%for 
during the optimisation process rather than just choosing one of them and following 
%it.
this choice.

Figures~\ref{fig:bar_plot_vehicle} and~\ref{fig:dot_plot_vehicle} present similar bar and dot plots for the Vehicle dataset. 
%when PNN was trained using the cHM algorithm. 
It can be seen that for this dataset 
%the Hybrid method 
cHM chose the FPA technique most often 
%for 
across all iterations. It 
%might illustrate 
suggests that when one particular swarm optimisation approach has some advantage over %others
the remaining ones, it is picked up more often by the 
%cHM 
algorithm.

\section{Conclusions}
\label{sec:Conclusions}

%Tables \ref{table:comparison_avg_test_accuracy} - \ref{table:comparison_avg_test_recall} 
The results show that the constrained Hybrid Metaheuristic method overperforms single metaheuristics in the 
%Probabilistic Neural Network 
PNN training for the classification task. 
%The data introduced 
It is shown that 
%the 
cHM is capable of selecting 
%the best 
a suitable metaheuristic for a given 
%part of the problem at a time
stage of the training process. 
%It usually selects the most optimal of all metaheuristics and helps to find the most suitable one for a given problem for further evaluation.

In addition, cHM shortens the time needed to evaluate the metaheuristics.
%of $N$. 
In fact, applying CHM 
%it 
is roughly $N$ times faster than testing each of the $N$ metaheuristic separately, with the simplifying assumption that all methods have the same training time 
%constraint.
requirements.

%Figures \ref{fig:bar_plot_cancer} and \ref{fig:dot_plot_cancer} 
Furthermore, the results show 
%the distribution of the selection of inside optimisers by the cHM algorithm for the Cancer dataset. It can be seen 
that 
%the Hybrid 
cHM 
%method 
picks the inside optimizers effectively from a set of available 
%different 
swarm methods during the overall optimisation process, and 
%for 
in specific iterations. 
%It establishes 
The above observations lay the foundation for the cHM ability 
%of the technique 
to combine various weak methods into a coherent stronger optimiser.

In 
%further 
future work, the transmission of the population and metaheuristic parameters between metaheuristics 
%should 
could be studied in more detail. In addition, the sensitivity of 
%the 
cHM to probing time 
%should 
could be tested to find the optimal 
%value 
range of the probing / fit time trade-off. In the end, performing the HPO of the cHM method for each dataset might lead to further improvements in PNN training performance with this algorithm.

\section*{Acknowledgments}
Jacek Ma{\'n}dziuk was supported by the National Science Centre, grant number 2023/49/B/ST6/01404 and partially by funding from the Polish Ministry of Science and Higher Education assigned to the AGH University of Krakow. The research project was partially supported by the program „Excellence Initiative – research university” for the AGH University of Krakow and was partially supported by a Grant for Statutory Activity from the Faculty of Physics and Applied Computer Science of the AGH.
 \bibliographystyle{elsarticle-num} 
 \bibliography{bibliografia}





% \end{thebibliography}
\end{document}